\documentclass[10pt,conference,letterpaper]{IEEEtran}
\usepackage{kotex}
\IEEEoverridecommandlockouts
\usepackage{cite}
\usepackage{amsmath,amssymb,amsfonts}
\usepackage{algorithmic}
\usepackage{graphicx}
\usepackage{textcomp}
\usepackage{xcolor}

\usepackage{subcaption}
\usepackage{url}
\usepackage{booktabs}
\usepackage{multirow}
\usepackage{tabularx}

\usepackage{xcolor}
\definecolor{takeawaybg}{gray}{0.96}
\definecolor{takeawayframe}{gray}{0.55}
\newcommand{\takeawaybox}[1]{%
\vspace{0.45em}%
\noindent%
\begingroup%
\setlength{\fboxsep}{3.5pt}%
\setlength{\fboxrule}{0.35pt}%
\fcolorbox{takeawayframe}{takeawaybg}{%
\begin{minipage}{\dimexpr\columnwidth-2\fboxsep-2\fboxrule\relax}
\textbf{Takeaway.} #1
\end{minipage}%
}%
\endgroup%
\vspace{0.45em}%
}

\usepackage{pifont}
\newcommand{\omark}{\ensuremath{\bigcirc}}
\newcommand{\xmark}{\ding{55}}
\usepackage{makecell}
\usepackage{adjustbox}
\usepackage{enumitem}
\usepackage{eso-pic}

\def\BibTeX{{\rm B\kern-.05em{\sc i\kern-.025em b}\kern-.08em
    T\kern-.1667em\lower.7ex\hbox{E}\kern-.125emX}}
\begin{document}
\bstctlcite{IEEEexample:BSTcontrol}

\title{Not All AI Agents Are Equal: Characterizing Resource and Performance Dynamics}
\author{%
  \IEEEauthorblockN{
    Wonmi Choi$^{1,*}$,
    Minuk Park$^{1}$,
    Zhixiong Niu$^{2}$,
    Yongqiang Xiong$^{2}$,
    Chuck Yoo$^{1,\dagger}$,
    Gyeongsik Yang$^{1,\dagger}$
  }
  \IEEEauthorblockA{
    $^1$Department of Computer Science and Engineering, Korea University \quad
    $^2$Microsoft Research Asia\\
    % \{ymcui, mupark\}@os.korea.ac.kr,
    % \{zhniu, yqx\}@microsoft.com,
    % chuckyoo@os.korea.ac.kr,
    % g\_yang@korea.ac.kr
  }
  \thanks{$^*$Work partially done during an internship at Microsoft Research Asia.}%\vspace{-1em}
  \thanks{$^\dagger$Corresponding authors.}\vspace{-1em}
  
}

\AddToShipoutPicture* {
  \AtPageLowerLeft {
    \put(0, 40){
      \makebox[\paperwidth][c]{
        \begin{minipage}{\textwidth}
          \centering
          \footnotesize
          \copyright~2026 IEEE. Personal use of this material is permitted. Permission from IEEE must be obtained for all other uses, in any current or future media, including reprinting/republishing this material for advertising or promotional purposes, creating new collective works, for resale or redistribution to servers or lists, or reuse of any copyrighted component of this work in other works. \\
          This paper has been accepted for publication in IEEE MASCOTS 2026. 
        \end{minipage}
      }
    }
  }
}

\maketitle

\begin{abstract}
LLM-based AI agents process user requests through iterative reasoning and tool execution, often involving the invocation of remote LLM APIs with local tool containers. This execution model can make the optimization of agent serving difficult because latency, local resource demand, and container bottlenecks inter-mix across requests. However, the current agent ecosystem runs without much consideration of resource dynamics, which results in significant waste of the precious resources. This paper analyzes the resource inter-mix of AI agents for three representative tasks: retrieval-augmented question answering, web search, and software coding. To this end, we characterize the latency with respect to the resource dynamics of processing multiple requests and tasks concurrently. Our measurements show that agents have a wide range of behaviors depending on tasks, so that even the same tool can differ substantially in resource dynamics. We also find that running multiple requests concurrently exposes task-dependent bottlenecks in resource dynamics such as CPU, disk I/O, and memory. Furthermore, we uncover that faster LLM responses or more CPU cores do not always accelerate agents. Based on these observations, we demonstrate new optimization opportunities that exploit the resource dynamics of tasks: CPU-aware tool admission and task-aware CPU allocation. Our results show that the latency of CPU-sensitive agent tasks improves $\sim$5.4$\times$, and the average latency across multiple tasks is reduced $\sim$32\% compared to native agents.

\end{abstract}


\section{Introduction}
Large language model (LLM)-based AI agents automate user requests through iterative reasoning and tool execution. A single agent request is completed over multiple iterations, each consisting of an LLM-based reasoning step followed by a tool execution step that runs external tools according to the reasoning result. This workflow enables diverse tasks such as retrieval-augmented question answering (RAQA), web search, and coding (software engineering), but also makes agent serving different from conventional single-turn LLM serving: latency and resource demand are determined by multiple heterogeneous iterations rather than by a single inference.

A common deployment pattern of AI agents is to separate LLM inference from local tool execution. Because hosting a competitive LLM locally requires substantial hardware resources~\cite{pan2025cost}, and closed-source models often achieve strong accuracy on agentic tasks~\cite{swebench_leaderboard,bfcl_leaderboard}, agent services commonly rely on remote LLM APIs, such as Gemini, GPT, and Claude. In contrast, tool execution commonly runs locally in containers (also denoted as sandboxes), as tools can consume resources, access external services, and mutate execution state.\footnote{Some AI agents run tool containers in the cloud, but the tool runtime is still separated from the remote LLM inference service.} This pattern is common in practical agents, such as Codex CLI, Claude Code, and GitHub Copilot agent mode, which interact with remote LLMs while reading files, editing code, running commands, and executing tests in local~\cite{openai_codex_cli, claude, github_copilot}.

A key challenge in serving agents is understanding the performance and resource usage of tool containers. A single agent request alternates among remote LLM inference, external Web/API calls, and local tool execution. Since tool execution directly contributes to end-to-end latency, the resource demand and bottlenecks of tool containers can significantly affect the agent performance \cite{raj2025cpu}. This makes system-level optimizations such as resource allocation and scheduling important\cite{go2026making, yoo2025revisiting, choi2024intelligent}. However, such optimization is difficult because agents do not follow a fixed tool execution path. Unlike conventional microservices, where each request is mostly processed inside a service container, an agent request invokes local tools only at certain steps and waits for remote services at others. As a result, a local container may remain idle during remote waiting periods, but later require bursty CPU or large resident memory during tool execution. Under-provisioning delays the latency-critical tool execution, whereas static over-provisioning wastes resources during remote-wait phases.

Prior work has begun to study LLM-based agents from several angles, including CPU-centric execution~\cite{raj2025cpu}, infrastructure cost~\cite{kim2026cost}, and OS-level resource control for agent containers~\cite{zheng2026agentcgroup}. However, they offer only partial views: none jointly characterizes how task type, request concurrency, remote LLM response time, and tool-container CPU allocation shape latency and local resource usage. This leaves it unclear whether an agent is limited by CPU contention, resident memory footprint, disk I/O pressure, or remote LLM waiting.

This paper presents a comprehensive analysis of AI agents across three representative tasks---RAQA, web search, and software coding. We organize our analysis around four questions: (1) how latency is divided across LLM API calls, external Web/API calls, and local tool execution; (2) which local resources become bottlenecks under concurrent requests; (3) how LLM response time changes the overlap of local tool executions; and (4) how tool-container CPU allocation affects latency and resource efficiency.

Our measurements lead to three main observations. First, agent latency and local resource demand are not tightly coupled. For example, RAQA is dominated by local retrieval and resident index memory, whereas web search is slow mostly due to remote Web/API waits with little local CPU use. This heterogeneity suggests that agent-serving systems should not apply the same optimization across tasks without understanding the tool's behavior. Second, the concurrency exposes different bottlenecks across tasks, including CPU contention in local-computation-heavy tasks, disk I/O pressure in write-heavy coding tasks, and memory growth from shared or per-request state. Thus, scalable agent serving requires bottleneck-aware resource orchestration rather than CPU- or memory-only control. Third, LLM response time and CPU allocation interact in a non-trivial way: faster LLM responses can increase the overlap of local tool executions, while additional CPU cores help only when local execution is on the critical path. Therefore, simply using a faster LLM or assigning more CPU does not always accelerate agent requests.

Furthermore, we show that these observations can guide optimization opportunities for agent serving: observation-guided CPU-aware tool admission and task-aware CPU allocation. Experiment results show that our analysis-guided optimization improves latency for CPU-sensitive agent tasks by up to 5.4$\times$ and reduces the mixed-task average latency by up to 32\%.
In short, we make the following contributions.
\begin{itemize}[leftmargin=1em,topsep=0pt,nosep,itemsep=0pt,parsep=0pt,partopsep=0pt]
\item \textbf{Characterization.} Characterize latency and local resource usage across three agent task types and seven benchmarks.
\item \textbf{Bottleneck analysis.} Identify how CPU, memory, and disk I/O bottlenecks emerge under concurrent requests.
\item \textbf{Sensitivity analysis.} Analyze how LLM response time and tool-container CPU allocation jointly affect latency and resource contention.
\item \textbf{Optimization opportunities.} Show that observation-guided decisions achieve up to 5.4$\times$ CPU-sensitive task speedup and 32\% mixed-workload latency reduction.
\end{itemize}

\section{Background}

\subsection{AI Agent Workflow}\label{subsec:agent_workflow}
A task in AI agents specifies the type of work the agent performs and the tools available for that work. Representative tasks include RAQA, web search, and coding. A workflow defines how the agent performs a task by processing each user request through LLM and diverse tools.
%invocation, tool execution, request memory state updates, and repeated iterations until producing the final answer.
ReAct~\cite{yao2023react} is a representative workflow for LLM-based AI agents. In this workflow, an agent orchestrator maintains a request memory state for each user request, storing the request, intermediate reasoning results, tool outputs, and other context required across iterations. Each iteration consists of two steps: reasoning step and tool execution step.

In the reasoning step, the agent invokes the LLM with the current request memory state. The LLM then determines whether the request can be completed and generates either a final answer or one or more tool calls. If the LLM generates tool calls, the agent proceeds to the tool execution step. In the tool execution step, the agent executes the requested tools and collects their outputs. When multiple independent tool calls are generated in the same iteration, they can be executed in parallel. The resulting tool outputs are then incorporated into the request memory state and used as input to the next iteration's reasoning step. This process repeats until the LLM generates the final answer.\footnote{Other agent workflows may differ in LLM invocation frequency, planning/reflection logic, or tool-level parallelism, but many follow the same high-level loop of reasoning, tool execution, and state update as ReAct.}

\begin{figure}[]
    \centering
    \captionsetup[subfigure]{justification=centering,singlelinecheck=false}
    \includegraphics[width=\linewidth]{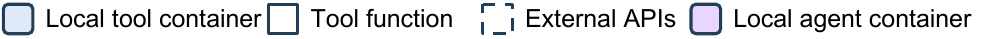}   
    \vspace{0.1em}
    \begin{subfigure}[t]{0.32\linewidth}
        \centering
        \includegraphics[height=3.5cm]{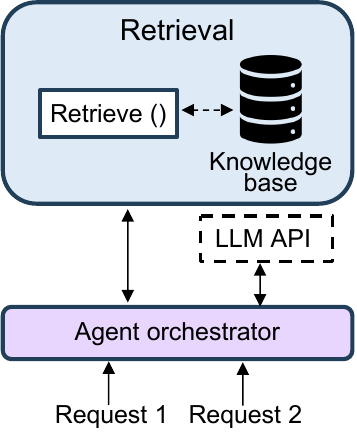}
        \caption{RAQA}
        % \caption{Retrieval-\\augmented QA}
        \label{fig:rag}
    \end{subfigure}\hfill
    \begin{subfigure}[t]{0.32\linewidth}
        \centering
        \includegraphics[height=3.5cm]{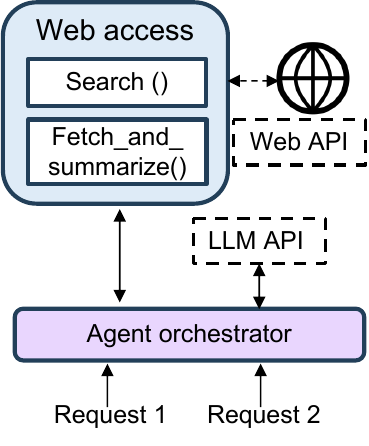}
        \caption{Web search}
        \label{fig:web_search}
    \end{subfigure}\hfill
    \begin{subfigure}[t]{0.32\linewidth}
        \centering
        \includegraphics[height=3.5cm]{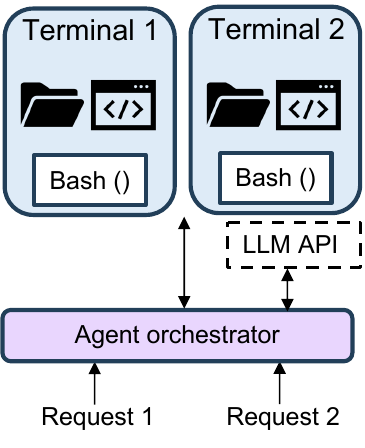}
        \caption{Coding}
        \label{fig:coding}
    \end{subfigure}\vspace{-.2em}
    \caption{Container structures of representative tasks.}
    \label{fig:tool_workloads}\vspace{-1.8em}
\end{figure}

\subsection{Container Structures of Agent Tools}

In many agent systems, the reasoning step is conducted by remote LLM APIs, while the tool execution step runs locally in separate runtimes. This separation means that agent latency is not determined only by remote LLM response time; it also depends on how local tool executions consume CPU, memory, and I/O resources on the serving infrastructure.

Tools are commonly implemented as containers. For example, agent tools include shell commands, file editing, database queries, Web/API access, test execution, and application-specific code. As they consume CPU and memory, access external services, and mutate filesystem or process state, agent systems run them in isolated containers rather than directly on the host. Recent platforms also reflect this deployment model (e.g., GKE Sandbox and Cloudflare Sandbox).

The container structure differs across task categories. Fig. \ref{fig:tool_workloads} illustrates the structure of tool containers for three representative agent tasks: RAQA, web search, and coding. Each example shows two concurrent user requests for the task, denoted as Request 1 and Request 2. The blue boxes represent tool containers, which run the tools required for each task and isolate tool execution. The white boxes represent callable tool functions exposed to the LLM. The dashed boxes indicate external services, such as remote LLM services or Web APIs, that are accessed over the network but are not directly managed by the agent framework. The purple boxes represent agent orchestration containers, which receive user requests, coordinate LLM calls and tool invocations, and maintain per-request memory state. Our analysis focuses on tool containers and external APIs, as the orchestrator in general does not create noticeable bottlenecks.

The container structure differs in how local tool runtimes are shared across requests. RAQA (Fig. \ref{fig:rag}) and web search (Fig. \ref{fig:web_search}) both use a shared tool container: a single tool container handles tool invocations from multiple concurrent requests. In RAQA, this container hosts the knowledge base and exposes \texttt{retrieve()} tool that returns top-ranked evidence passages for a natural-language query. In web search, the shared container exposes Web access tools, such as \texttt{search()} and \texttt{fetch\_and\_summarize()}, which call external Web APIs and return search results or summarized page contents. Thus, although RAQA mainly relies on a shared local retrieval state and web search additionally depends on external web services, both follow the same shared-container structure for local tool execution across multiple requests.

In contrast, coding tasks (Fig. \ref{fig:coding}) use request-specific tool containers. Each request runs its own terminal container that exposes \texttt{bash()} tool, allowing the agent to execute commands, run tests, and modify files within an isolated filesystem and process environment. Therefore, coding agents commonly use separate containers for different requests, such as Terminal~1 and Terminal~2 in the figure, to avoid interference and preserve request-specific execution state.
In the following subsection, we review existing studies on AI agents and explain the need for our comprehensive analysis.

\section{Related Work and Research Gap}

\begin{table}[t]
\centering
\caption{Related work comparison.}
\label{tab:related_work}\vspace{-.3em}
\footnotesize
\setlength{\tabcolsep}{3.0pt}
\renewcommand{\arraystretch}{0.4}
\begin{adjustbox}{max width=\columnwidth}
\begin{tabular}{@{}cccccc@{}}
\toprule
\multicolumn{2}{c}{\textbf{Study}}
& \textbf{Ours}
& \makecell[c]{\textbf{Raj \cite{raj2025cpu}}}
& \makecell[c]{\textbf{Kim \cite{kim2026cost}}}
& \makecell[c]{\textbf{AgentCgroup \cite{zheng2026agentcgroup}}} \\
\midrule

\multicolumn{2}{c}{\textbf{Runtime type}}
& Containers
& \makecell[c]{Host\\process}
& \makecell[c]{Host\\process}
& Containers \\
\midrule

\multirow{3}{*}{\makecell[c]{\textbf{Analyzed}\\\textbf{tasks}}}
& \makecell[c]{Retrieval-\\augmented QA} 
& \omark & \omark & \xmark & \xmark \\
\cmidrule(lr){2-6}
& Web search 
& \omark & \omark & \omark & \xmark \\
\cmidrule(lr){2-6}
& Coding 
& \omark & \omark & \omark & \omark \\
\midrule

\multirow{3}{*}{\makecell[c]{\textbf{Per-task}\\\textbf{analysis}}}
& CPU    
& \omark & \xmark & \xmark & \omark \\
\cmidrule(lr){2-6}
& Memory 
& \omark & \xmark & \xmark & \omark \\
\cmidrule(lr){2-6}
& GPU    
& \xmark & \xmark & \omark & \xmark \\
\midrule

\multirow{3}{*}{\makecell[c]{\textbf{Concurrency}\\\textbf{analysis}}}
& CPU    
& \omark & \xmark\textsuperscript{*} & \xmark & \xmark \\
\cmidrule(lr){2-6}
& Memory 
& \omark & \xmark & \xmark & \xmark \\
\cmidrule(lr){2-6}
& GPU    
& \xmark & \xmark\textsuperscript{*} & \omark & \xmark \\
\midrule

\multirow{2}{*}{\makecell[c]{\textbf{Sensitivity}\\\textbf{analysis}}}
& LLM latency    
& \omark & \omark & \xmark & \xmark \\
\cmidrule(lr){2-6}
& CPU allocation 
& \omark & \xmark & \xmark & \xmark \\
\bottomrule
\end{tabular}
\end{adjustbox}

\begin{minipage}{\columnwidth}
\vspace{.2em}
\footnotesize
\xmark\textsuperscript{*}: measures throughput under batched requests, not resource usage directly.
\end{minipage}
\vspace{-2.5em}
\end{table}

We review existing studies on AI agents and identify the remaining gaps that motivate our analysis.
Several prior studies analyzed AI agents, as summarized in Table~\ref{tab:related_work}. We compare them by runtime type, task coverage, and analysis scope, including per-task resource usage, concurrency behavior, and sensitivity to key agent conditions such as remote LLM latency and container CPU allocation.

First, existing studies differ in runtime type and task coverage. Raj et al.\cite{raj2025cpu} analyzed RAQA, web search, and coding agents, and Kim et al.\cite{kim2026cost} studied web search and coding agents. However, both studies evaluated agents running as host processes. This setting does not capture the resource behavior of containerized agents, where tool execution is isolated and constrained by container-level resource controls. AgentCgroup~\cite{zheng2026agentcgroup} studied containerized agents, but focused only on coding tasks. Thus, existing studies did not characterize containerized agent execution across diverse task types. Our work fills this gap by analyzing three representative tasks: RAQA, web search, and coding.

Second, in terms of per-task analysis, existing studies provided only partial evidence on agent bottlenecks and sometimes attributed the dominant bottleneck to different resources. Raj et al. \cite{raj2025cpu} identified CPU as a major bottleneck, based mainly on throughput measurements rather than direct per-task resource usage measurements. AgentCgroup\cite{zheng2026agentcgroup} measured CPU and memory, but only for coding tasks, and reported memory as the dominant bottleneck. This suggests that agent bottlenecks can vary, but prior work lacks a comprehensive analysis across both resources and task types. We address this gap by characterizing CPU and memory usage across RAQA, web search, and coding, providing a more systematic understanding of agents. We exclude GPU utilization from our analysis as LLM inference is via external APIs, such as Claude or GPT, whose actual GPU usage is not visible.

Third, in terms of concurrency analysis, existing studies provided only limited insight into how resource bottlenecks evolved under concurrent requests for each task. This analysis is important because practical agent services often handle multiple user requests simultaneously \cite{agentix}, and their local tool executions can overlap even when each request waits for remote LLM or Web/API responses at different times. 
AgentCgroup \cite{zheng2026agentcgroup} estimated CPU and memory usage for 32--64 concurrent requests from profiles of a single request resource, without directly conducting the experiments. Raj et al.~\cite{raj2025cpu} discussed possible CPU and GPU bottlenecks mainly through throughput measurements, rather than direct resource-usage measurements. In contrast, we vary the request rate and directly analyze CPU and memory usage across RAQA, web search, and coding. This allows us to show not only which resource becomes the bottleneck for each task, but also how the bottleneck changes and affects end-to-end latency as concurrency increases.

Lastly, existing studies provided limited analysis of sensitivity to LLM response latency and container CPU allocation. LLM response latency varies across inference models; even for the same model, it can also change over time due to queueing delays or network interference in the model-serving backend~\cite{artificial_analysis_provider_benchmark}. Container CPU allocation is also important because tools run inside containers, and their execution speed is bounded by the assigned CPU resources. Raj et al.~\cite{raj2025cpu} studied LLM latency effects by assuming LLM inference on two GPU types. In contrast, we jointly vary LLM response latency and container CPU allocation to analyze how the two together affect bottlenecks, latency, and resource-use efficiency.

\section{Comprehensive Analysis on AI Agents}\label{analysis}
This section presents our comprehensive analysis of AI agents. We first describe the experiment setup and analysis methodology, and then report our findings.

\begin{table}[t]
\centering
\caption{Benchmarks and the tools in analysis.}
\label{tab:bench}\vspace{-.3em}
\footnotesize
\setlength{\tabcolsep}{3.0pt}
\renewcommand{\arraystretch}{0.5}
\begin{tabularx}{\columnwidth}{@{}l>{\raggedright\arraybackslash}p{3.6cm}>{\raggedright\arraybackslash}X@{}}
\toprule
\textbf{Task type} & \textbf{Benchmarks} & \textbf{Tools} \\
\midrule

RAQA
& HotpotQA~\cite{yang2018hotpotqa},
  TriviaQA~\cite{joshi2017triviaqa}
& Hybrid retrieval~\cite{sawarkar2024blended}\newline
  (BM25, FAISS~\cite{douze2024faiss}) \\

\midrule

Web search
& BFCLv4~\cite{patil2025bfcl},
  QASC~\cite{khot2020qasc}
& Tavily API~\cite{tavily},
  LexRank\newline summarization~\cite{erkan2004lexrank} \\

\midrule

Coding
& SWE-bench~\cite{jimenez2024swebench},
  BigCode-\newline Bench~\cite{zhuo2025bigcodebench},
  APPS~\cite{hendrycks2021apps}
& Bash shell \\

\bottomrule
\end{tabularx}
\vspace{-2em}
\end{table}

\begin{table*}[t]
\centering
\caption{Analysis dimensions and goals.}
\label{tab:analysis_plan}\vspace{-.6em}
% \footnotesize
\setlength{\tabcolsep}{2pt}
\renewcommand{\arraystretch}{0.48}
\begin{adjustbox}{max width=\textwidth}
\begin{tabular}{@{}llll@{}}
\toprule
\textbf{Analysis} & \textbf{Variable} & \textbf{Metrics} & \textbf{Goal} \\
\midrule
Task characterization (\S\ref{analysis:task:characterization})
& Task type (with sequential request)
& Latency breakdown, CPU, memory
& Identify local execution fraction from latency and resource usage \\
% & Identify local execution fraction from end-to-end and resource usage per task \\
\midrule
Concurrent request analysis (\S\ref{analysis:concurrent:characterization})
& Request rate: 0.5--6 req/s
& Latency, CPU, memory
& Identify resource bottlenecks under concurrent requests \\
\midrule
Sensitivity analysis (\S\ref{analysis:sensitivity})
& LLM response time, CPU allocation
& Latency, CPU throttling ratio
& Analyze sensitivity of remote LLM waiting and CPU capacity\\
\bottomrule
\end{tabular}
\end{adjustbox}
\vspace{-1.9em}
\end{table*}

\subsection{Setup and Methodology}
\subsubsection{Benchmarks and tools} We evaluate agents on seven benchmarks for three types of agent tasks: RAQA, web search, and coding (Table \ref{tab:bench}). For each task type, we equip the agent with the tools that the task commonly requires. For RAQA, we use HotpotQA \cite{yang2018hotpotqa} and TriviaQA \cite{joshi2017triviaqa} benchmarks, which require retrieving supporting evidence to answer multi-hop or open-domain questions over a Wikipedia passage corpus, following common RAG settings \cite{sawarkar2024blended, yu2024rankrag}. For web search, we use BFCLv4 \cite{patil2025bfcl} and QASC \cite{khot2020qasc}, which require agents to gather web or domain-specific facts and synthesize evidence, reflecting current agent behavior \cite{openai_codex_cli, claude, gemini}. For coding, we use SWE-bench \cite{jimenez2024swebench}, BigCodeBench \cite{zhuo2025bigcodebench}, and APPS \cite{hendrycks2021apps}, covering repository-level issue resolution, practical Python code generation with library calls, and algorithmic programming problems.
 We run all experiments on an Azure VM with 24 vCPUs (AMD EPYC 7V13) and 216 GB memory.

\subsubsection{LLM API} For each reasoning step of a request, the agent queries the Gemini 3 Flash API \cite{gemini}. LLMs are known to produce non-deterministic outputs \cite{artificial_analysis_provider_benchmark, yuan2025nondeterminism}. If left uncontrolled, this variability would confound our runtime measurements and make comparisons across configurations unreliable. To ensure a fair and consistent evaluation, we first run each benchmark sequentially and record the agent trajectory, including LLM outputs and per-step latency. In the remaining experiments, we replay the recorded LLM outputs and execute the corresponding tool calls.

\subsubsection{Analysis items}
We analyze AI agents along three dimensions, as summarized in Table~\ref{tab:analysis_plan}. First, for task characterization (\S\ref{analysis:task:characterization}), we run requests sequentially and measure latency breakdown, CPU usage, and memory usage to identify each task's local execution behavior. Second, for concurrent execution, we vary the request rate from 0.5 to 6 requests per second (req/s) and measure how latency and resource usage change. When CPU and memory do not explain latency growth, we further measure disk write throughput and CPU I/O wait ratio to identify I/O bottlenecks. Third, for sensitivity analysis (\S\ref{analysis:sensitivity}), we scale the recorded LLM response times by 1$\times$, 2$\times$, and 3$\times$, vary tool-container CPU allocation from 2 to 8 cores, and measure end-to-end latency and CPU throttling ratio. CPU and memory usage are collected from container cgroups and reported as both average and peak values.

\begin{figure*}[t]
\centering
\captionsetup[subfigure]{font=footnotesize,skip=2pt}
\begin{subfigure}[t]{0.248\textwidth}
    \centering
    \includegraphics[width=\linewidth]{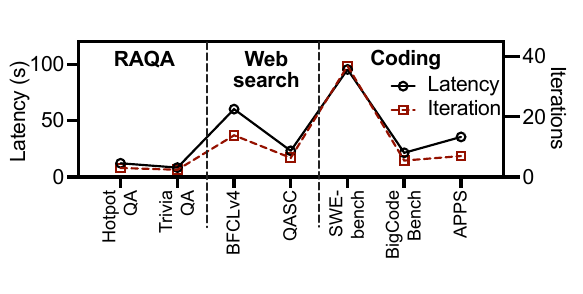}
    \caption{End-to-end latency and iterations}
    \label{fig:e2e}
\end{subfigure}\hfill
\begin{subfigure}[t]{0.248\textwidth}
    \centering
    \includegraphics[width=\linewidth]{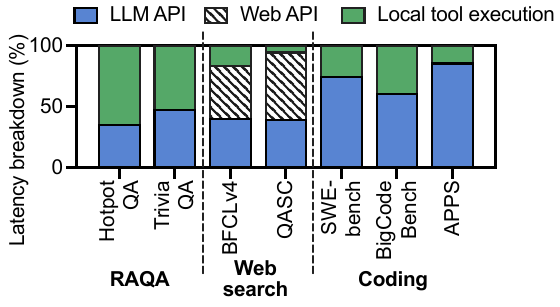}
    \caption{Latency breakdown}
    \label{fig:latency_breakdown}
\end{subfigure}\hfill
\begin{subfigure}[t]{0.248\textwidth}
    \centering
    \includegraphics[width=\linewidth]{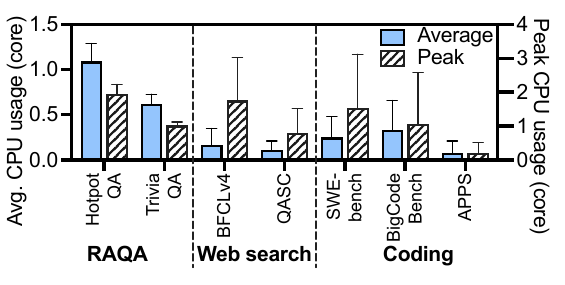}
    \caption{CPU usage}
    \label{fig:cpu_usage}
\end{subfigure}\hfill
\begin{subfigure}[t]{0.248\textwidth}
    \centering
    \includegraphics[width=\linewidth]{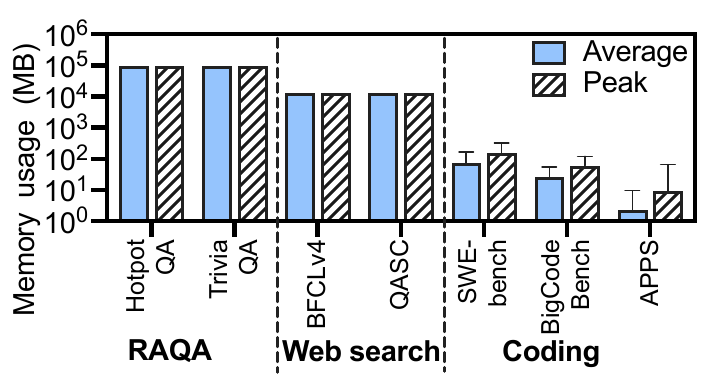}
    \caption{Memory usage}
    \label{fig:mem_usage}
\end{subfigure}
\vspace{-0.2em}
\caption{Task characterization analysis (\S\ref{analysis:task:characterization}).}
\label{fig:task_characterization}
\vspace{-1.8em}
\end{figure*}

\subsection{Task Characterization}\label{analysis:task:characterization}
We characterize each task by analyzing where its end-to-end latency is spent and which local resources it consumes under sequential request execution.

\subsubsection{End-to-end latency}
Fig. \ref{fig:e2e} reports the average end-to-end latency (left y-axis) and the number of iterations (right y-axis) per request across the seven benchmarks (x-axis). Latency scales with the iteration count: each iteration runs one LLM API call and triggers tool execution, so tasks that require more iterations cost more.

Among the three task types, RAQA requires the fewest iterations, with 3.0 for HotpotQA and 2.4 for TriviaQA on average. This is because a single retrieval step returns candidate passages, allowing the agent to produce an answer in only a few steps and resulting in low latency. Web search requires substantially more iterations, with 13.9 for BFCLv4 and 6.4 for QASC, because the agent first issues a URL search and then fetches and reads multiple pages before synthesizing an answer. This increases latency by up to 7$\times$. Coding shows the largest variation, ranging from 7 to 36.9 iterations and from 21.5 to 95.4 s of latency, depending on how much code the agent must inspect and modify. For example, BigCodeBench typically requires implementing a single function from a specification, whereas SWE-bench requires exploring a repository, locating relevant files, and editing them. We next break down this latency in detail.

\subsubsection{Latency breakdown}
Fig. \ref{fig:latency_breakdown} splits end-to-end latency into LLM API time, Web API time, and tool execution time. The dominant component differs by task. RAQA is dominated by tool execution time for local retrieval, accounting for 62\% of the average latency across HotpotQA and TriviaQA. Web search is dominated by Web API time, 49\% on average across BFCLv4 and QASC, while its tool execution on local result summarization is only 10.5\%. Coding is dominated by LLM API time; tool execution for local Bash execution accounts for 39\% on average across the three coding benchmarks.
The point is that long end-to-end latency does not always mean heavy local tool use: a task can be slow mostly while waiting for remote LLM or Web API responses by diverse tools. So, local container provisioning should account for the local tool execution rather than end-to-end latency.

\subsubsection{CPU usage}
Fig. \ref{fig:cpu_usage} shows the average and peak CPU usage, where bars indicate the average and whiskers denote the standard deviation across requests of each benchmark. Average CPU usage is measured over the duration of each request, including both local tool execution and waiting for LLM or Web API responses; peak CPU usage is the maximum demand during local tool execution. Average CPU usage (blue bars) follows the local tool execution fraction in Fig. \ref{fig:latency_breakdown}, as a request consumes CPUs during local execution and remains idle while waiting for remote responses. RAQA, whose local retrieval is the largest share of its latency (62\%), has the highest average CPU usage (86\%), while web search, dominated by remote Web API time with only 10.5\% local, has the lowest (14\%); coding (39\% local execution) lies in between (22\%).

Peak CPU usage (hatched bars) reflects how each task runs tools in parallel.
Even within the same task type, peak CPU demand differs across benchmarks. For example, in RAQA, the difference between HotpotQA and TriviaQA comes from parallel retrieval: as explained in \S\ref{subsec:agent_workflow}, independent tool calls within an iteration run in parallel. So, HotpotQA, which issues 1.94 concurrent tool calls on average, peaks near 2 cores, whereas TriviaQA, which issues 1.01 concurrent tool calls on average, peaks near 1 core. Other task types also show higher peak CPU when they execute more tools in parallel.

We also analyze the standard deviation, which shows the variation across requests within each task. RAQA tasks show low variance (0.19 cores on average), because each retrieval call has nearly constant cost: the index search uses a fixed-size corpus and returns a fixed number of passages. In contrast, web search shows bigger variation (0.98 cores on average), as its peak CPU usage depends on the amount of text fetched to summarize. The size of fetched text varies by more than three orders of magnitude across requests, leading to large variation even among requests from the same benchmark. For coding tasks, different benchmarks show different trends. For example, SWE-bench and BigCodeBench show the highest variances (1.57 and 1.49 cores). In SWE-bench, each request runs a different repository-level test suite, so the cost of its tool calls varies widely across requests. Also, BigCodeBench requests are highly diverse, ranging from lightweight utility functions to CPU-intensive executions with large inputs or expensive tests. In contrast, APPS shows low variation (0.29 cores), as each request involves only one or two Python programs for algorithmic problem solving.

\subsubsection{Memory usage}
Fig.~\ref{fig:mem_usage} shows mean and peak memory usage. Mean usage is measured over each request's execution, while peak usage is the highest memory mesured during the execution. Whiskers show the standard deviation across all requests in each benchmark. In RAQA and web search tasks, a single tool container stays resident across requests, keeping retrieval indices and loaded libraries in memory throughout; each request's footprint is therefore roughly flat, so average and peak memory nearly coincide (within 1\%). In coding tasks, memory usage varies across benchmarks, as each request runs different tools in a fresh container that starts with minimal memory, with memory dynamically allocated during execution and released afterward. As a result, peak memory exceeds average memory by 2--4$\times$. Across benchmarks, RAQA uses the most memory, up to 98 GB on average, as the retriever keeps its index resident in memory. Web search uses about 13 GB, mainly from loaded libraries. Coding tasks use far less memory but vary widely, from 2 to 73 MB on average, depending on each task's runtime and libraries. Peak memory follows the same ordering: 99 GB for RAQA, 13 GB for web search, and 9--156 MB for coding.

In short, three tasks show distinct resource usage. RAQA spends most of its latency in local retrieval, sustaining high, steady CPU usage and a large resident index in memory. Web search spends most of its latency waiting on remote APIs, so its CPU usage is low and varies with fetched text, with resident memory. Coding spends most of its latency in LLM API calls, making highly variable CPU and memory usage per request.

The analysis leads to two observations. First, end-to-end latency does not reflect tool resource demand directly. For example, web search is among the slowest tasks yet uses the least CPU, as its latency is dominated by remote waiting. Second, each task stresses a different resource, so no single resource assumption fits all agent tasks.

\takeawaybox{End-to-end latency does not directly reflect tool resource demand: each task and its tools stress a different resource. Managing resources for agents thus requires knowing which tools run and how often, not just latency.}

\begin{figure}[t]
    \centering
    \captionsetup[subfigure]{font=footnotesize,skip=2pt}

    \begin{subfigure}[t]{0.42\linewidth}
        \centering
        \includegraphics[width=\linewidth]{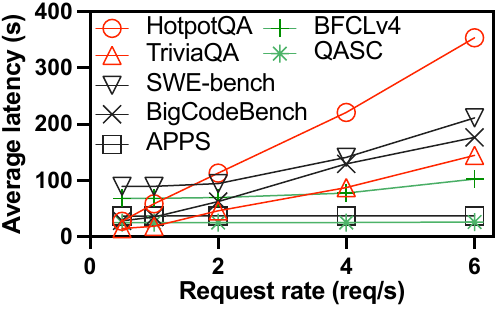}
        \caption{Latency}
        \label{fig:con_latency}
    \end{subfigure}
    \hfil
    \begin{subfigure}[t]{0.42\linewidth}
        \centering
        \includegraphics[width=\linewidth]{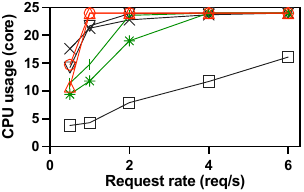}
        \caption{CPU usage}
        \label{fig:con_cpu}
    \end{subfigure}

    \vspace*{-.5em}

    \begin{subfigure}[t]{0.5\linewidth}
        \centering
        \includegraphics[width=\linewidth]{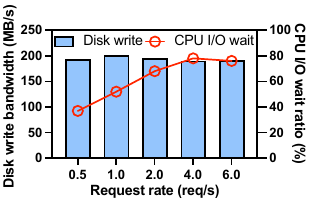}
        \caption{Disk write \& I/O wait}
        \label{fig:con_io}
    \end{subfigure}
    \hfil
    \begin{subfigure}[t]{0.46\linewidth}
        \centering
        \includegraphics[width=\linewidth]{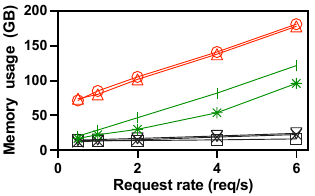}
        \caption{Memory usage}
        \label{fig:con_memory}
    \end{subfigure}
    \vspace{-0.3em}
    \caption{Concurrent request analysis (\S\ref{analysis:concurrent:characterization}).}
    \label{fig:concurrency}
    \vspace{-1.9em}
\end{figure}

\subsection{Concurrent Request Analysis}\label{analysis:concurrent:characterization}
We run concurrent requests for each task at rates from 0.5 to 6 req/s and measure how end-to-end latency, CPU usage, and memory usage scale. Although we omit the full results due to space constraints, we empirically verify that LLM API latency remains stable across the different request rates. Therefore, following prior studies~\cite{feng2025agentrr}, we replay recorded LLM trajectory: we reuse the LLM outputs and response latencies recorded during sequential execution while executing the corresponding tool calls under concurrent requests.

\subsubsection{Latency}

Fig.~\ref{fig:con_latency} shows average latency as the request rate increases. When the request rate increases by 12$\times$, from 0.5 to 6 req/s, latency grows very differently across tasks. RAQA increases the most: HotpotQA rises by 13.1$\times$ and TriviaQA by 10.2$\times$. In contrast, web search changes only slightly, with BFCLv4 increasing by 1.5$\times$ and QASC by 1.04$\times$. Coding lies between the two but varies by benchmark: BigCodeBench rises by 6.4$\times$, SWE-bench by 2.4$\times$, and APPS shows no noticeable change.

This latency ordering follows the local tool execution share in the latency breakdown (Fig. \ref{fig:latency_breakdown}). Local tool execution contends for local container resources under concurrent requests, whereas remote LLM and Web API waits can overlap without consuming local CPU. RAQA, where local retrieval accounts for 62\% of latency, is therefore the most affected. Web search, where local summarization accounts for only 10.5\% and most of its latency is spent on Web APIs, is nearly unaffected. Coding has a local execution share of 39\%, but its sensitivity depends on the tools each benchmark runs per request.

\subsubsection{CPU usage}
Fig.~\ref{fig:con_cpu} shows peak CPU usage as the request rate increases. We report peak rather than average usage because CPU contention appears during local execution bursts: a task can become CPU-bound when its burst demand saturates the cores, even if its average CPU usage remains low due to remote-wait phases. We therefore identify the CPU-bound point at which peak CPU usage reaches the entire core capacity and no longer increases.
RAQA becomes CPU-bound earliest, at 1 req/s, so its latency increase mainly comes from CPU contention in local retrieval. Web search saturates later, at 2 req/s for BFCLv4 and 4 req/s for QASC, and its latency increases less because local summarization accounts for only a small fraction of its end-to-end latency. 

Coding tasks show different trends between benchmarks: SWE-bench becomes CPU-bound at 2 req/s, matching its latency increase, while APPS never saturates, matching its flat latency above.
In terms of BigCodeBench, although its latency increases substantially with the request rate, it becomes CPU-bound only at 4 req/s, so CPU contention alone cannot explain its latency increase. To identify the cause, we examine disk write throughput and the CPU I/O wait ratio of the benchmark in Fig. \ref{fig:con_io}. Even at the lowest rate of 0.5 req/s, disk write throughput already reaches about 200 MB/s, close to the 207 MB/s ceiling measured with \texttt{fio}, which indicates that the disk is saturated from the beginning. As the request rate increases, the CPU I/O wait ratio also increases, meaning that CPU cores spend more time waiting for disk writes to complete. Disk I/O, rather than CPU, is therefore the primary bottleneck for BigCodeBench under concurrency.

\subsubsection{Memory usage}
Fig. \ref{fig:con_memory} shows memory usage under different request rates. As the request rate increases, memory usage grows by 2.5$\times$ on average for RAQA, 5.8$\times$ for web search, and 1.6$\times$ for coding. 
RAQA has the highest memory usage across all request rates because the tasks maintain large in-memory retrieval indices. This creates a high memory footprint even at 0.5 req/s. Increasing the request rate adds intermediate per-request state on top of this shared index. 

For web search, the memory usage itself is lower than RAQA's, but the amount of increase from 0.5 to 6 req/s is relatively similar between the two tasks (near 100 GB). So, web search starts with a smaller memory than RAQA, but the memory state required per request is similar for both.
Coding shows the smallest memory growth. Although each request runs in its own container, most execution state is written to files rather than kept as long-lived in-memory objects. As a result, the per-request resident memory footprint remains relatively small even as the number of concurrent requests increases.

In short, concurrency does not uniformly stress a single resource. CPU contention limits compute-heavy tasks such as RAQA and SWE-bench, disk I/O limits the write-heavy coding benchmark BigCodeBench, and memory usage increases with request rate in RAQA and web search.

\takeawaybox{Concurrency shows task-dependent bottlenecks across CPU, disk I/O, and memory. This could complement prior CPU- or memory-focused techniques for agents \cite{raj2025cpu, zheng2026agentcgroup} and motivate task-aware optimization, such as task-aware admission control and resource allocation.}

\begin{figure*}[t]
    \centering
    \captionsetup[subfigure]{font=footnotesize,skip=1pt}

    \begin{subfigure}[t]{0.28\textwidth}
        \centering
        \includegraphics[width=\linewidth]{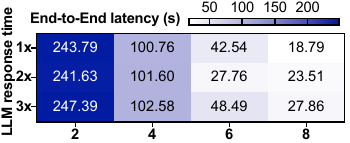}
        \caption{RAQA, end-to-end latency}
        \label{fig:heatmap_latency_hotpotqa}
    \end{subfigure}
    \hfil
    \begin{subfigure}[t]{0.28\textwidth}
        \centering
        \includegraphics[width=\linewidth]{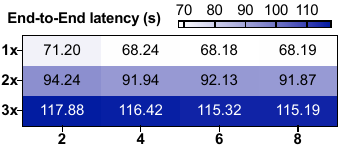}
        \caption{Web search, end-to-end latency}
        \label{fig:heatmap_latency_bfcl}
    \end{subfigure}
    \hfil
    \begin{subfigure}[t]{0.28\textwidth}
        \centering
        \includegraphics[width=\linewidth]{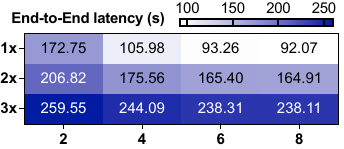}
        \caption{Coding, end-to-end latency}
        \label{fig:heatmap_latency_swebench}
    \end{subfigure}

    % \vspace{-0.3em}

    \begin{subfigure}[t]{0.28\textwidth}
        \centering
        \includegraphics[width=\linewidth]{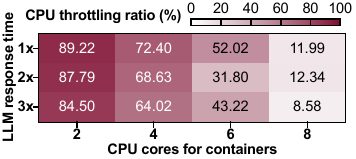}
        \caption{RAQA, CPU throttling ratio}
        \label{fig:heatmap_throttle_hotpotqa}
    \end{subfigure}
    \hfil
    \begin{subfigure}[t]{0.28\textwidth}
        \centering
        \includegraphics[width=\linewidth]{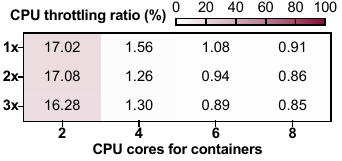}
        \caption{Web search, CPU throttling ratio}
        \label{fig:heatmap_throttle_bfcl}
    \end{subfigure}
    \hfil
    \begin{subfigure}[t]{0.28\textwidth}
        \centering
        \includegraphics[width=\linewidth]{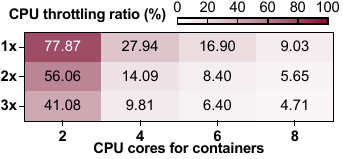}
        \caption{Coding, CPU throttling ratio}
        \label{fig:heatmap_throttle_swebench}
    \end{subfigure}
    \vspace{-0.3em}
    \caption{Sensitivity analysis (\S\ref{analysis:sensitivity}): x-axis of CPU cores given for containers and y-axis of LLM response time.}%Sensitivity of request latency and CPU throttling to CPU allocation and LLM response latency.}
    \label{fig:latency_throttle_heatmap}
    \vspace{-1.5em}
\end{figure*}

\subsection{Sensitivity Analysis}\label{analysis:sensitivity}

We jointly vary LLM response time and tool-container CPU allocation to see how they shape end-to-end latency. Due to space constraints, the remaining experiments use one representative benchmark per task type: HotpotQA for RAQA, BFCLv4 for web search, and SWE-bench for coding. At a fixed request rate of 0.5 req/s, we simultaneously vary two factors: LLM response time, scaled by 1$\times$, 2$\times$, and 3$\times$ relative to the time measured under sequential request execution, and the total CPU allocation for tool containers, set from 2 to 8 cores. We report two metrics: 1) mean end-to-end latency across requests and 2) CPU throttling ratio, defined as the fraction of scheduling periods in which a container is throttled at its CPU limit, as an indicator of CPU contention.

Fig.~\ref{fig:latency_throttle_heatmap} shows the results, with the LLM response time on the y-axis and the tool-container CPU allocation on the x-axis. Figs. \ref{fig:heatmap_latency_hotpotqa}--\ref{fig:heatmap_latency_swebench} report the mean end-to-end latency, where darker blue indicates higher latency; Figs. \ref{fig:heatmap_throttle_hotpotqa}--\ref{fig:heatmap_throttle_swebench} report the CPU throttling ratio, where deeper red indicates more throttling.

For RAQA, end-to-end latency in Fig.~\ref{fig:heatmap_latency_hotpotqa} is much more sensitive to CPU allocation (x-axis) than to LLM response time (y-axis). When the CPU allocation increases from 2 to 8 cores at the same LLM response time, end-to-end latency decreases by 13.0$\times$, 10.3$\times$, and 8.9$\times$ for the 1$\times$, 2$\times$, and 3$\times$ settings, respectively. In contrast, when the LLM response time increases from 1$\times$ to 3$\times$ at the same CPU allocation, latency increases only by 1.01$\times$ at 2 cores, 1.02$\times$ at 4 cores, and 1.14$\times$ at 6 cores; the effect becomes larger only at 8 cores, where latency increases by 1.48$\times$ because CPU throttling is already mostly relieved. The throttling ratio (Fig. \ref{fig:heatmap_throttle_hotpotqa}) shows the same pattern. Increasing the CPU allocation from 2 to 8 cores reduces throttling by 7.4$\times$, 7.3$\times$, and 9.3$\times$ across the three LLM response time values, whereas increasing the LLM response time from 1$\times$ to 3$\times$ at the same CPU allocation changes throttling by at most 1.3$\times$. This indicates that HotpotQA is dominated by CPU contention in local retrieval: additional CPU cores directly reduce throttling and latency, whereas slower LLM responses have only a limited effect while the task remains CPU-bound.

For web search, end-to-end latency (Fig.~\ref{fig:heatmap_latency_bfcl}) shows the opposite behavior: it is sensitive to LLM response time (y-axis) but almost insensitive to CPU allocation (x-axis). When the CPU allocation increases from 2 to 8 cores at the same LLM response time, latency decreases by only 1.04$\times$, 1.03$\times$, and 1.02$\times$ for the 1$\times$, 2$\times$, and 3$\times$ settings. In contrast, when the LLM response time increases from 1$\times$ to 3$\times$ at the same CPU allocation, end-to-end latency increases by about 1.7$\times$ across CPU allocations. The throttling ratio (Fig.~\ref{fig:heatmap_throttle_bfcl}) also shows that CPU contention is not a major bottleneck for BFCLv4: throttling is noticeable only at 2 cores and remains near zero from 4 cores onward. This indicates that BFCLv4 is dominated by remote waiting rather than local CPU contention: additional CPU cores provide little end-to-end latency benefit, whereas slower LLM responses directly increase latency.

For coding, end-to-end latency (Fig.~\ref{fig:heatmap_latency_swebench}) is affected by both CPU allocation (x-axis) and LLM response time (y-axis), but the dominant factor changes across settings. When the CPU allocation increases from 2 to 8 cores at the same LLM response time, latency decreases by 1.88$\times$ for the 1$\times$ setting, but only by 1.25$\times$ and 1.09$\times$ for the 2$\times$ and 3$\times$ settings, respectively. Thus, additional CPU cores are most beneficial when LLM responses are fast. In contrast, when the LLM response time increases from 1$\times$ to 3$\times$ at the same CPU allocation, latency increases by 1.50$\times$ at 2 cores, but by 2.30$\times$, 2.55$\times$, and 2.59$\times$ at 4, 6, and 8 cores. The throttling ratio (Fig. \ref{fig:heatmap_throttle_swebench}) explains this behavior. At 2 cores, increasing the LLM response time from 1$\times$ to 3$\times$ reduces throttling by 1.9$\times$, from 78\% to 41\%, because slower LLM responses spread out when requests enter local tool execution and partially relieve CPU contention. This indicates that SWE-bench lies between HotpotQA and BFCLv4: additional CPU cores help when tool executions overlap heavily, but slower LLM responses reduce this overlap and lower the marginal benefit of extra CPU cores.

Overall, the three tasks represent different trends. RAQA (HotpotQA) is local-CPU-bound, so end-to-end latency mainly follows CPU allocation. Web search (BFCLv4) is remote-wait-dominated, so latency mainly follows LLM response time. Coding (SWE-bench) is mixed: LLM response time changes how much tool execution overlaps across requests, which in turn changes the benefit of extra cores.

\takeawaybox{Sensitivity to LLM response time and CPU allocation varies across tasks: depending on the task, agents can tolerate slower LLM responses or fewer CPU cores with only a little additional latency, suggesting that operators can avoid unnecessary CPU over-provisioning by identifying the more sensitive factor for each task.}

\begin{figure}[t]
    \centering
    \captionsetup{justification=centering}

    % --- Block 1: CPU-aware tool admission control latency ---
    \begin{minipage}[t]{0.23\textwidth}
        \centering
        \captionsetup[subfigure]{justification=centering,singlelinecheck=false}
        \includegraphics[width=0.75\linewidth]{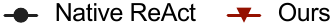}   
        \vspace{0.1em}
        
        \begin{subfigure}[t]{0.48\linewidth}
            \centering
            \includegraphics[height=1.85cm]{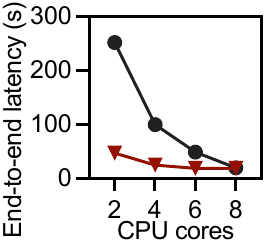}
            \caption{HotpotQA}
            \label{fig:quota_hotpotqa}
        \end{subfigure}
        % \hfill
        \begin{subfigure}[t]{0.48\linewidth}
            \centering
            \includegraphics[height=1.85cm]{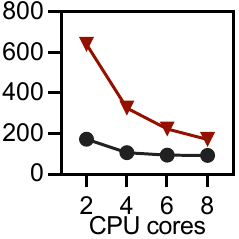}
            \caption{SWE-bench}
            \label{fig:quota_swe-bench}
        \end{subfigure}
        \caption{CPU-aware tool admission (\S\ref{subsec:cpu_aware_tool_admission}).}
        \label{fig:quota}
    \end{minipage}%
    % --- Block 3: task-aware CPU allocation ---
    \begin{minipage}[t]{0.25\textwidth}
        \centering
        \captionsetup[subfigure]{justification=centering,singlelinecheck=false}
        \includegraphics[width=\linewidth]{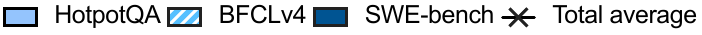}
        \par\vspace{0.3em}
        \begin{subfigure}[b]{0.49\linewidth}
            \centering
            \includegraphics[height=1.73cm]{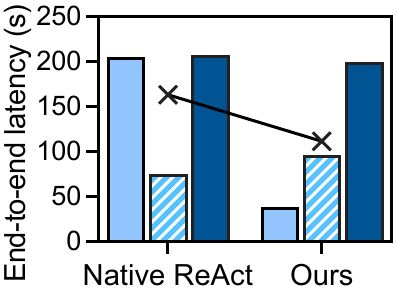}
            \caption{6-CPU}
            \label{fig:usecase_6cpu}
        \end{subfigure}\hfill
        \begin{subfigure}[b]{0.49\linewidth}
            \centering
            \includegraphics[height=1.73cm]{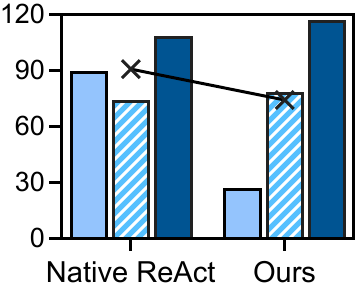}
            \caption{12-CPU}
            \label{fig:usecase_12cpu}
        \end{subfigure}
        \caption{Task-aware CPU allocation (\S\ref{subsec:task_aware_cpu_allocation}).}
        \label{fig:usecase}
    \end{minipage}%
    \vspace{-1.7em}
\end{figure}

\section{Analysis-Guided Optimization Opportunities}
Here, we demonstrate how the analysis and observations in \S\ref{analysis} can translate into practical optimization opportunities. Our goal is not to introduce a complete technique, but to show that decisions guided by our analysis can improve end-to-end latency and efficiency. We present two cases: CPU-aware tool admission and task-aware CPU allocation.

\subsection{Case 1: CPU-aware Tool Admission}
\label{subsec:cpu_aware_tool_admission}

The sensitivity analysis in \S\ref{analysis:sensitivity} suggests that reducing CPU contention can improve end-to-end latency only for tasks whose latency is strongly sensitive to CPU allocation. HotpotQA (RAQA) shows the largest latency reduction as CPU allocation increases, indicating that it is strongly CPU contention-bound. This suggests an optimization opportunity: instead of increasing CPU cores, CPU contention can be reduced by controlling how many tool executions run simultaneously. As changing the CPU cores on running containers is challenging, we consider this approach.

However, this approach is not suitable for every task. BFCLv4 (web search) shows little CPU sensitivity and less than 20\% CPU throttling across 2–8 cores (Fig.~\ref{fig:latency_throttle_heatmap}), so limiting tool executions would mostly add queueing delay without meaningfully reducing CPU contention. SWE-bench (coding) shows mixed sensitivity: CPU allocation affects latency, but the benefit is smaller and less consistent than in HotpotQA. Therefore, limiting concurrent tool executions is expected to increase SWE-bench latency because the added queueing delay in waiting can outweigh the smaller CPU-side gain.

Based on this observation, we design and test \emph{CPU-aware tool admission} policy at the same request rate as in \S\ref{analysis:sensitivity}, 0.5 req/s. The policy limits the number of concurrently running tool executions to the number of allocated CPU cores. For example, with two CPU cores, at most two tool executions run at once, while the remaining executions wait in a queue. We compare this policy with the native ReAct workflow, which starts all tool executions without admission control. We evaluate HotpotQA and SWE-bench, but the analysis suggests different outcomes for the two: the policy should improve HotpotQA, while it should increase latency for SWE-bench.\footnote{We omit BFCLv4 due to space constraints because the analysis (\S\ref{analysis:sensitivity}) indicates that it is not a CPU-contention-bound task; under admission control, its latency also does not improve, similar to SWE-bench.}

Fig. \ref{fig:quota} shows the end-to-end latency. The results match the expectation from the analysis. For HotpotQA, our policy reduces latency by 3.23$\times$ on average (Fig. \ref{fig:quota_hotpotqa}), confirming that limiting concurrent tool executions can improve latency when a task is strongly CPU-contention-bound. In contrast, for SWE-bench, the same policy rather increases latency by 2.76$\times$ on average (Fig.~\ref{fig:quota_swe-bench}).

The results show that CPU-aware tool admission should be applied in a task-adaptive manner. It is effective for CPU-contention-bound tasks such as RAQA, whose latency strongly decreases with additional CPU cores. However, it is unsuitable for tasks such as coding, where reducing tool concurrency can significantly increase queueing delays.

\subsection{Case 2: Task-Aware CPU Allocation}
\label{subsec:task_aware_cpu_allocation}
We next examine whether task-dependent CPU sensitivity from \S\ref{analysis:sensitivity} can improve mixed-agent workloads. We mix three benchmarks that represent different task types: HotpotQA for RAQA, BFCLv4 for web search, and SWE-bench for coding. Each task type arrives at 0.5 req/s, resulting in an aggregate rate of 1.5 req/s.

We design and test \emph{task-aware CPU allocation} policy that assigns different numbers of CPU cores to different tasks. We compare it with native ReAct-based baseline, where all tasks follow ReAct workflow and receive equal CPU allocation. In contrast, our policy allocates cores based on per-task CPU sensitivity. We evaluate two total CPU budgets: 6 and 12 cores. Based on the sensitivity analysis in \S\ref{analysis:sensitivity}, we allocate more cores to the CPU-sensitive RAQA benchmark, HotpotQA, and fewer cores to the CPU-insensitive web search benchmark, BFCLv4. For SWE-bench, which shows mixed sensitivity, we keep the allocation close to the uniform allocation. With the 6-core budget, BFCLv4, HotpotQA, and SWE-bench receive 1, 3, and 2 cores, respectively, compared with 2 cores each in native ReAct. With the 12-core budget, they receive 2, 6, and 4 cores, compared with 4 cores each in native ReAct. We also apply the CPU-aware tool admission from \S\ref{subsec:cpu_aware_tool_admission} to HotpotQA because it reduces CPU throttling for this benchmark.

Fig.~\ref{fig:usecase} compares native ReAct and our policy in terms of average end-to-end latency. Bars show the latency of each benchmark, and lines with x markers show the total average latency across all benchmarks. For the 6-core budget (Fig. \ref{fig:usecase_6cpu}), our policy reduces the total average latency by 32\% compared with native ReAct. This improvement mainly comes from HotpotQA, whose latency decreases by 5.4$\times$ by receiving one additional core and applying CPU-aware tool admission. Although BFCLv4 latency increases by 28\% due to the decreased CPU, its impact on the total average latency is smaller than the HotpotQA improvement. 

For the 12-core budget (Fig.~\ref{fig:usecase_12cpu}), our policy reduces the total average latency by 23\% compared with native ReAct. HotpotQA latency decreases by 3.2$\times$ with additional cores and CPU-aware tool admission, while BFCLv4 sees little latency increase despite receiving fewer cores. The results suggest the potential of CPU sensitivity-guided resource orchestration, where cores can be shifted from CPU-insensitive to CPU-sensitive tasks with limited latency penalty.

\section{Conclusion and Future Work}
This paper characterizes LLM-based AI agents across task type, concurrency, LLM response time, and tool-container CPU allocation. Our analysis shows that agents are not homogeneous: RAQA is highly sensitive to local CPU contention, web search is dominated by remote-service waiting, and coding exhibits mixed bottlenecks from local commands, disk I/O, and request-specific runtime state.
We further show that these observations can guide optimization opportunities. CPU-aware tool admission and task-aware CPU allocation improve latency by reducing unnecessary CPU contention and shifting cores toward CPU-sensitive tasks. As future work, we plan to generalize the ideas into online resource manager for heterogeneous tasks and multi-agent deployments.

\section*{Acknowledgment}
This research was supported by Microsoft Research Asia, by the Ministry of Science, ICT (MSIT), Korea, under the Global Research Support Program in the Digital Field program (RS-2024-00436680) supervised by the IITP, by the Basic Science Research Program through the National Research Foundation of Korea funded by the Ministry of Education (MOE) (RS-2021-NR060143), by an NRF grant funded by the Korea government (MSIT) (RS-2024-00336564), and by the ICT Creative Consilience Program through IITP grant funded by the MSIT (IITP-2026-RS-2020-II201819).

\bibliographystyle{IEEEtran}
\bibliography{reference}

\end{document}